\documentclass{article}
\usepackage{icml2007} 
\usepackage{epsf}
\usepackage{mlapa}

\icmltitlerunning{A Neural Network Approach to Ordinal Regression}

\begin{document} 

\twocolumn[
\icmltitle{A Neural Network Approach to Ordinal Regression}

\icmlauthor{}{}
%\icmladdress{\textbf{Ordinal regression, neural network, perceptron, ranking}}
% The following author list should only appear in the accepted versions. 
\icmlauthor{Jianlin Cheng}{jcheng@cs.ucf.edu}
 \icmladdress{School of Electrical Engineering and Computer Science, 
             University of Central Florida, Orlando, FL 32816, USA}

\vskip 0.3in
]

\begin{abstract}
Ordinal regression is an important type of learning, which has properties of both classification
and regression. Here we describe a simple and effective approach to adapt a traditional neural network to 
learn ordinal categories. Our approach is a generalization of the perceptron method for ordinal regression. 
On several benchmark datasets, our method (NNRank) outperforms a 
neural network classification method. Compared with the ordinal regression methods 
using Gaussian processes and support vector machines, NNRank achieves comparable
performance. Moreover, NNRank has the advantages of traditional neural networks: learning
in both online and batch modes, handling very large training datasets, and making rapid predictions.    
These features make NNRank a useful and complementary tool for large-scale data processing tasks such as information
retrieval, web page ranking, collaborative filtering, and protein ranking in Bioinformatics.  \\

\end{abstract}

\section{Introduction}
Ordinal regression (or ranking learning) is an important supervised problem of 
learning a ranking or ordering on instances, which has the property of both 
classification and metric regression. The learning task of ordinal regression is to
assign data points into a set of finite ordered categories. For example, a 
teacher rates students' performance using A, B, C, D, and E 
(A $>$ B $>$ C $>$ D $>$ E) (Chu \& Ghahramani, 2005a).
Ordinal regression is different from classification due to the order of categories.
In contrast to metric regression, the response variables (categories) in 
ordinal regression is discrete and finite.

The research of ordinal regression
dated back to the ordinal statistics methods in 1980s (McCullagh, 1980; McCullagh \& Nelder, 1983) and 
machine learning research in 1990s (Caruana et al., 1996; Herbrich et al., 1998; Cohen et al., 1999).
It has attracted the considerable attention in recent years due to its
potential applications in many data-intensive domains such as information retrieval (Herbrich et al., 1998),
web page ranking (Joachims, 2002), collaborative filtering (Goldberg et al., 1992; Basilico \& Hofmann, 2004; Yu et al., 2006), 
image retrieval (Wu et al., 2003), and protein ranking (Cheng \& Baldi, 2006) in 
Bioinformatics. 

A number of machine learning methods have been developed or redesigned to address ordinal regression
problem (Rajaram et al., 2003), including 
perceptron (Crammer \& Singer, 2002) and its kernelized generalization (Basilico \& Hofmann, 2004),
neural network with gradient descent (Caruana et al., 1996; Burges et al., 2005),
Gaussian process (Chu \& Ghahramani, 2005b; Chu \& Ghahramani, 2005a; Schwaighofer et al., 2005),
large margin classifier (or support vector machine) (Herbrich et al., 1999; Herbrich et al., 2000; Joachims, 2002; Shashua \& Levin, 2003; Chu \& Keerthi, 2005; Aiolli \& Sperduti, 2004; Chu \& Keerthi, 2007),
k-partite classifier (Agarwal \& Roth, 2005),
boosting algorithm (Freund et al., 2003; Dekel et al., 2002),
constraint classification (Har-Peled et al., 2002), 
regression trees (Kramer et al., 2001),
Naive Bayes (Zhang et al., 2005),
Bayesian hierarchical experts (Paquet et al., 2005),
binary classification approach (Frank \& Hall, 2001; Li \& Lin, 2006) that decomposes the original
ordinal regression problem into a set of binary classifications,
and the optimization of nonsmooth cost functions (Burges et al., 2006).

Most of these methods can be roughly classified into two categories: 
pairwise constraint approach (Herbrich et al., 2000; Joachims, 2002; Dekel et al., 2004; Burges et al., 2005) and 
multi-threshold approach (Crammer \& Singer, 2002; Shashua \& Levin, 2003; Chu \& Ghahramani, 2005a).
The former is to convert the full ranking relation into pairwise order constraints. The latter 
tries to learn multiple thresholds to divide data into ordinal categories. 
Multi-threshold approaches 
also can be unified under the general, extended binary classification framework (Li \& Lin, 2006).

The ordinal regression methods have different advantages and disadvantages. 
Prank (Crammer \& Singer, 2002), a perceptron approach that
generalizes the binary perceptron algorithm to the ordinal multi-class situation, 
is a fast online algorithm. However, like a standard perceptron method, its accuracy suffers
when dealing with non-linear data, while a quadratic kernel version of Prank 
greatly relieves this problem. 
One class of accurate large-margin classifier approaches (Herbrich et al., 2000; Joachims, 2002)
convert the ordinal relations into $O(n^2)$ ($n$: the number of data points) 
pairwise ranking constraints for the structural 
risk minimization (Vapnik, 1995; Schoelkopf \& Smola, 2002). 
Thus, it can not be applied to medium size datasets 
($>$ 10,000 data points), without discarding some pairwise preference relations. 
It may also overfit noise due to incomparable pairs. 

The other class of powerful large-margin classifier 
methods (Shashua \& Levin, 2003; Chu \& Keerthi, 2005) generalize the support vector
formulation for ordinal regression by finding $K-1$ thresholds on the real line that divide  
data into $K$ ordered categories. The size of this optimization problem
is linear in the number of training examples. However, like support vector machine used for 
classification, the prediction speed
is slow when the solution is not sparse, which makes it not appropriate for
time-critical tasks. Similarly, another state-of-the-art approach, Gaussian process method (Chu \& Ghahramani, 2005a),  
also has the difficulty of handling large training datasets 
and the problem of slow prediction speed in some situations.
%due to its non-sparse solution.

Here we describe a new neural network approach for ordinal regression that 
has the advantages of 
neural network learning: learning in both online and batch mode, training on very large dataset (Burges et al., 2005), handling
non-linear data, good performance, and rapid prediction. Our method can 
be considered a generalization of the perceptron    
learning (Crammer \& Singer, 2002) into multi-layer perceptrons (neural network) for ordinal regression. 
Our method is also related to the classic generalized linear models (e.g.,
cumulative logit model) for ordinal regression (McCullagh, 1980).  
Unlike the neural network method (Burges et al., 2005) trained on pairs of  
examples to learn pairwise order relations, 
our method works on individual data points and uses multiple output nodes to
estimate the probabilities of ordinal categories. Thus, our method falls into the category of multi-threshold approach.
The learning of our method proceeds similarly as traditional neural networks using back-propagation (Rumelhart et al., 1986). 

On the same benchmark datasets, our method yields the performance better than the standard classification 
neural networks and comparable to the state-of-the-art methods using support vector machines and 
Gaussian processes. In addition, our method can learn on very large datasets and make rapid predictions.  

\section{Method}
\subsection{Formulation}
Let $D$ represent an ordinal regression dataset consisting of $n$ data points ($x,y$) 
, where $x \in R^d$ is an input feature vector
and $y$ is its ordinal category from a finite set $Y$.
Without loss of generality, we assume that 
$Y={1,2,...,K}$ with "$<$" as order relation.  

For a standard classification neural network without considering 
the order of categories, the goal is 
to predict the probability of a data point $x$ belonging to one 
category $k$ ($y=k$). The input is $x$ and the target of
encoding the category $k$ is a vector $t$ = $(0,...,0, 1, 0, ...,0)$,
where only the element $t_k$ is set to 1 and all others to 0. The goal
is to learn a function to map input vector $x$ to a probability 
distribution vector $o = (o_1,o_2,...o_k,...o_K)$, where $o_k$ is closer to 1  
and other elements are close to zero, subject to the constraint $\sum_{i=1}^{K} o_i = 1$.

In contrast, like the perceptron approach (Crammer \& Singer, 2002), our neural network approach considers 
the order of the categories.
If a data point $x$ belongs to category $k$, it is classified automatically 
into lower-order categories ($1,2,...,k-1$) as well.  
So the target vector of $x$ is $t = (1,1,..,1,0,0,0)$,
where $t_i\ (1 \leq i \leq k)$ is set to 1 and other elements zeros. Thus, the goal is
to learn a function to map the input vector $x$ to a probability
vector $o = (o_1,o_2,...,o_k,...o_K)$, where $o_i\ (i \leq k)$ is close to 1
and  $o_i\ (i \geq k)$ is close to 0. $\sum_{i=1}^{K} o_i$ is the estimate of 
number of categories (i.e. $k$) that $x$ belongs to, instead of 1.
The formulation of 
the target vector is similar to the perceptron approach (Crammer \& Singer, 2002). It is also 
related to  
the classical cumulative probit model for ordinal regression (McCullagh, 1980), in the sense
that we can consider the output probability vector $(o_1,...o_k,...o_K)$ as 
a cumulative probability distribution on categories $(1,...,k,...,K)$, 
i.e., $\frac{\sum_{i=1}^{K} o_i}{K}$ is the proportion of categories
that $x$ belongs to, starting from category 1. 

The target encoding scheme of our method is related to but, 
different from multi-label learning (Bishop, 1996)
and multiple label learning (Jin \& Ghahramani, 2003)
because our method imposes 
an order on the labels (or categories). 

\subsection{Learning}
\label{learning}
Under the formulation, we can use the almost exactly same neural network machinery
for ordinal regression.
We construct a multi-layer neural network to learn ordinal relations from
$D$. The neural network has $d$ inputs corresponding to the number of dimensions of 
input feature vector $x$ and $K$ output nodes corresponding to $K$
ordinal categories. 
There can be one or more hidden layers. Without loss
of generality, we use one hidden layer to construct a standard two-layer feedforward
neural network. Like a standard neural network for classification, 
input nodes are fully
connected with hidden nodes, which in turn are fully connected with 
output nodes. Likewise, the transfer function of hidden nodes can be 
linear function, sigmoid function, and tanh function that is used in our
experiment. The only difference from traditional neural network 
lies in the output layer. Traditional neural networks use softmax $\frac{e^{-z_i}}{\sum_{i=1}^{K} e^{-z_i}}$(or normalized 
exponential function) for output nodes, satisfying the constraint that
the sum of outputs $\sum_{i=1}^{K}o_i$ is 1. 
$z_i$ is 
the net input to the output node $O_i$. 

In contrast, each output node $O_i$ of our neural
network uses a standard sigmoid function $\frac{1}{1+e^{-z_i}}$, without including the outputs from other nodes. 
Output node $O_i$ is used
to estimate the probability $o_i$ that a data point belongs to category $i$
independently, without subjecting
to normalization as traditional neural networks do. 
Thus, for a data point $x$ of category $k$, the target
vector is $(1,,1,..,1,0,0,0)$, in which the first $k$ elements is 1 and others 0.
This sets the target value of output nodes $O_i$ ($i \leq k$)
to 1 and $O_i$ ($i>k$) to 0. 
The targets instruct the neural network to adjust weights to
produce probability outputs as close as possible to the target vector. 
It is worth pointing out that using independent sigmoid functions for output nodes
does not guaranteed the monotonic relation ($o_1 >= o_2 >= ... >= o_K$),
which is not necessary but, desirable for making predictions (Li \& Lin, 2006). 
A more sophisticated approach is to impose the inequality constraints on the outputs
to improve the performance. 
         
Training of the neural network for ordinal regression proceeds very similarly as standard neural networks.
The cost function for a data point $x$ can be relative entropy or square 
error between the target vector and the output vector. For relative entropy,
the cost function for output nodes is $f_c = \sum_{i=1}^{K}{(t_i\log o_i + (1-t_i)\log(1-o_i))}$. 
For square error, the 
error function is $f_c = \sum_{i=1}^{K}{ (t_i-o_i)^2 }$. 
Previous studies (Richard \& Lippman, 1991) on neural network cost functions show that 
relative entropy and square error functions usually 
yield very similar results. 
In our experiments, we use
square error function and standard back-propagation to train 
the neural network. 
The errors are propagated back to output nodes, and  
from output nodes to hidden nodes, and finally to input nodes.

Since the transfer function $f_t$ of output node $O_i$ is the independent 
sigmoid function $\frac{1}{1+e^{-z_i}}$, the derivative of $f_t$ of output node $O_i$ 
is $\frac{\partial f_t}{\partial z_i} = \frac{e^{-z_i}}{(1+e^{-z_i})^2}$ =
$\frac{1}{1+e^{-z_i}}(1-\frac{1}{1+e^{-z_i}})$ = $o_i(1 - o_i)$. 
Thus, the net error propagated to output node $O_i$ is $\frac{\partial f_c}{\partial o_i} \frac{\partial f_t}{\partial z_i}
= \frac{t_i-o_i}{o_i(1-o_i)} \times o_i(1-o_i) = t_i - o_i$ for relative entropy cost function,
$\frac{\partial f_c}{\partial o_i} \frac{\partial f_t}{\partial z_i} = -2(t_i - o_i) \times o_i(1-o_i) = -2o_i(t_i-o_i)(1-o_i)$
for square error cost function. 
The net errors are propagated through neural networks to adjust weights using gradient descent as traditional neural networks do.
%Since the derivative used to compute the net error of an output node does not involve the outputs of other nodes, 
%the computation of net errors of output nodes is
%simpler than standard classification neural networks using
%softmax transfer function. 

Despite the small difference in the transfer function and the computation of its derivative, 
the training of our method is the same as traditional neural networks. The network can be trained
on data in the online mode where weights are updated per example, or in the batch mode where
weights are updated per bunch of examples. 

\subsection{Prediction}
In the test phase, to make a prediction, our method scans output 
nodes in the order $O_1, O_2,...,O_K$. It stops when the output of a node
is smaller than the predefined threshold $T$ (e.g., 0.5) or no nodes left. 
The index $k$ of the last node $O_k$ whose output 
is bigger than $T$ is the predicted category of the data point.
%Another alternative is to scan output nodes from $O_K$ to $O_1$ 
%to find the first node $k$ whose output is bigger than $T$.
%Both approaches yield similar results. The former is used in our experiments.
%In theory, for ideal cases, $\sum_{i=1}^{K} o_i$ is also the reasonable 
%estimate of the target category $k$. 
 
\section{Experiments and Results}
\subsection{Benchmark Data and Evaluation Metric}
We use eight standard datasets for ordinal regression (Chu \& Ghahramani, 2005a) to benchmark our method. 
The eight datasets (Diabetes, Pyrimidines, Triazines, Machine CUP, Auto MPG,
Boston, Stocks Domain, and Abalone) are originally used for metric regression. Chu and Ghahramani (Chu \& Ghahramani, 2005a)
discretized the real-value targets into five equal intervals, corresponding to five 
ordinal categories. 
The authors randomly split each dataset into training/test datasets and repeated the partition 20 times independently.     
We use the exactly same partitions as in (Chu \& Ghahramnai, 2005a) to train and test our method. 
%An important trick for neural network learning is to normalize the feature values into 
%a comparable range such as [-1,1]. For a feature having values bigger than 1 or less than -1, we
%simply normalize all values of the feature by dividing them by the closest upper bound number 
%in [10, 100, 1000,...].   

We use the online mode to train neural
networks. The parameters to tune are the number 
of hidden units, the number of epochs, and the learning rate. We create a grid for these three
parameters, where the hidden unit number is in the range $[1..15]$, the epoch number in the set 
$(50, 200, 500, 1000)$,
and the initial learning rate in the range $[0.01..0.5]$.
During the training, the learning rate is halved if training errors 
continuously go up for a pre-defined number (40, 60, 80, or 100) of  epochs.
For experiments on each data split, the neural network parameters 
are {\it fully} optimized on the training data without
using any test data. 

%This effectively
%helps avoid the overfitting of the training data. 
%Each combination of parameter values 
%is used to train neural network models on the 90\% training data and validated on the remaining 10\%. 
%The best parameters are used to train a neural network model on the whole training data.
For each experiment, after the parameters are optimized on the training data, 
we train five models on the training data with the optimal parameters,
starting from different initial weights.
The ensemble of five trained models are then used to estimate 
the generalized performance on the test data. That is, the average output of 
five neural network models is used to make predictions.  

%Training is very fast on a Pentium IV Linux desktop computer. Table \ref{train-time} reports
%the training time on the nine datasets. It shows it takes only a couple seconds for 1000
%epochs training on most datasets. 
%The longest 
%training is 20 seconds on Abalone dataset which has 1000 training data points.   

%\begin{table}
%\caption{Training time on the nine datasets. Number of epochs is 1000. Time is measured in seconds.}
%\label{train-time}
%\vspace{0.3cm}
%{
%\scriptsize
%\tiny
%\begin{tabular}{llllllllll}
%\hline
%    & Diabetes & Pyrimidines & Triazines & Wisconsin & Machine & Auto & Boston & Stocks & Abalone \\ 
%\hline
%Time(s) &2 & 4 & 5 & 5 & 4 & 6 & 6 & 15 & 20 \\
%\hline
%\end{tabular}
%}
%\end{table}

We evaluate our method using zero-one error and mean absolute error as in (Chu \& Ghahramani, 2005a). 
Zero-one error is the percentage of
wrong assignments of ordinal categories. 
Mean absolute error is the root mean square difference 
between assigned categories ($k'$) and true categories ($k$) of all data points. 
For each dataset, the training and evaluation process is repeated 20 times on 20 data splits.
Thus, we compute the average error and the standard deviation of the two metrics as in (Chu \& Ghahramani, 2005a).

\subsection{Comparison with Neural Network Classification}
We first compare our method (NNRank) with a standard neural network classification method (NNClass). 
We implement both NNRank and NNClass using C++. NNRank and NNClass share most code with 
minor difference in the transfer function of output nodes and its derivative computation
as described in Section \ref{learning}.

\begin{table*}[h]
\caption{The results of NNRank and NNClass on the eight datasets. 
The results are the average error over
20 trials along with the standard deviation.}
\label{rank-class}
\begin{center}
{\footnotesize
\begin{tabular}{|c|c|c|c|c|}
\hline
 & \multicolumn{2}{c|}{Mean zero-one error} & \multicolumn{2}{c|}{Mean absolute error}\\
\hline
Dataset & NNRank & NNClass & NNRank & NNClass \\ \hline
Stocks & 12.68$\pm$1.8\% & 16.97$\pm$ 2.3\% & 0.127$\pm$0.01  & 0.173$\pm$0.02\\
Pyrimidines & 37.71$\pm$8.1\% & 41.87$\pm$7.9\% & 0.450$\pm$0.09 & 0.508$\pm$0.11 \\
Auto MPG & 27.13$\pm$2.0\% & 28.82$\pm$2.7\% & 0.281$\pm$0.02 & 0.307$\pm$0.03  \\
%Wisconsin & 64.06$\pm$4.4\% & 64.20$\pm$0.95\% & 1.08$\pm$0.094 &1.3602$\pm$0.0440 \\ 
Machine & 17.03$\pm$4.2\% & 17.80$\pm$4.4\% &0.186$\pm$0.04& 0.192$\pm$0.06 \\
Abalone & 21.39$\pm$0.3\% & 21.74$\pm$ 0.4\% & 0.226$\pm$0.01  & 0.232$\pm$0.01 \\ 
Triazines & 52.55$\pm$5.0\% & 52.84$\pm$5.9\% & 0.730$\pm$0.06 & 0.790$\pm$0.09 \\
Boston & 26.38$\pm$3.0\% & 26.62$\pm$2.7\% & 0.295$\pm$0.03 & 0.297$\pm$0.03 \\
Diabetes & 44.90$\pm$12.5\% & 43.84$\pm$10.0\%  & 0.546$\pm$0.15 & 0.592$\pm$0.09  \\
\hline
\end{tabular}
}
\end{center}
\end{table*}

As Table \ref{rank-class} shows, NNRank outperforms NNClass in all but one case 
in terms of both the mean-zero error and the mean absolute error. 
And on some datasets 
the improvement of NNRank over NNClass is sizable. For instance, 
on the Stock and Pyrimidines datasets, the mean zero-one error of NNRank is about 4\% less 
than NNClass; on four datasets (Stock, Pyrimidines, Triazines, and Diabetes) the
mean absolute error is reduced by about .05.     
The results show that the ordinal regression neural network 
consistently achieves the better performance
than the standard classification neural network.
To futher verify the effectiveness of the neural network ordinal regression approach, 
we are currently evaluating NNRank and NNclass 
on very large ordinal regression datasets in the bioinformatics domain (work in progress).

\subsection{Comparison with Gaussian Processes and Support Vector Machines}
To further evaluate the performance of our method, we compare NNRank with two Gaussian
process methods (GP-MAP and GP-EP) (Chu \& Ghahramani, 2005a) and 
a support vector machine method (SVM) (Shashua \& Levin, 2003) implemented in (Chu \& Ghahramani, 2005a).
The results of the three methods are quoted from (Chu \& Ghahramani, 2005a).
Table \ref{mean-zero-one} reports the zero-one error on the eight datasets.
%On the three datasets (Diabetes, Triazines,
%and Abalone), 
NNRank achieves the best results on Diabetes, Triazines, and Abalone,  
%On Diabetes dataset, the performance of NNRank
%is significantly better. 
GP-EP on Pyrimidines, Auto MPG, and Boston, 
GP-MAP on Machine, and SVM on Stocks. 
% respectively.
%On the other five datasets (Auto MPG, Boston, and Stocks),
%the performance of NNRank is at most a few percentage lower than the best method.  

Table \ref{absolute-error} reports the mean absolute error on the eight datasets.
NNRank yields the best results on Diabetes and Abalone, GP-EP on Pyrimidines, Auto MPG, 
and Boston, GP-MAP on Triazines and Machine, SVM on Stocks. 

%Since the accuracy differences between NNRank and other three methods (SVM, GP-MAP, and GP-EP) are within two standard deviations,
%we conducted paired t-tests on the zero-one errors (and the mean absolute errors)
%across eight datasets to compare their performance. 
%The pairwise t-tests shows that the difference between NNRank and other three methods
%is not significant.  
%Thus, on the eight datasets, the performance of NNRank is comparable
In summary, on the eight datasets, the performance of NNRank is comparable
to the three state-of-the-art methods for ordinal regression.  
%It is also worth noting that the accuracy differences between these methods are mostly
%within the range of two standard deviations.  

\begin{table*}
\caption{Zero-one error of NNRank, SVM, GP-MAP, and GP-EP on the eight datasets. 
SVM denotes the support vector machine method (Shashua \& Levin, 2003; Chu \& Ghahramani, 2005a). 
GP-MAP and GP-EP are two Gaussian process methods using Laplace approximation (MacKay, 1992) and 
expectation propagation (Minka, 2001) 
respectively (Chu \& Ghahramani, 2005a).
The results are the average error over
20 trials along with the standard deviation.
We use boldface to denote the best results.}
\label{mean-zero-one}
\begin{center}
{\footnotesize

\begin{tabular}{|c|c|c|c|c|}
\hline
Data & NNRank & SVM & GP-MAP & GP-EP \\ \hline
Triazines & \textbf{52.55$\pm$5.0\%}    & 54.19$\pm$1.5\% & 52.91$\pm$2.2\% & 52.62$\pm$2.7\% \\
Pyrimidines & 37.71$\pm$8.1\%  & 41.46$\pm$8.5\% & 39.79$\pm$7.2\% & \textbf{36.46$\pm$6.5\%} \\
Diabetes & \textbf{44.90$\pm$12.5\%}  & 57.31$\pm$12.1\% & 54.23$\pm$13.8\% & 54.23$\pm$13.8\% \\
%Wisconsin &  \textbf{64.06$\pm$4.4\%}  & 70.78$\pm$3.73\% & 65.00$\pm$4.71\% & 65.16$\pm$4.65\% \\
Machine &  17.03$\pm$4.2\%   & 17.37$\pm$3.6\% & \textbf{16.53$\pm$3.6\%} & 16.78$\pm$3.9\% \\
Auto MPG & 27.13$\pm$2.0\%   & 25.73$\pm$2.2\% & 23.78$\pm$1.9\% & \textbf{23.75$\pm$1.7\%} \\
Boston &  26.38$\pm$3.0\%  & 25.56$\pm$2.0\% & 24.88$\pm$2.0\% & \textbf{24.49$\pm$1.9\%} \\
Stocks & 12.68$\pm$1.8\%   & \textbf{10.81$\pm$1.7\%} & 11.99$\pm$2.3\% & 12.00$\pm$2.1\% \\
Abalone & \textbf{21.39$\pm$0.3\%}   & 21.58$\pm$0.3\% & 21.50$\pm$0.2\% & 21.56$\pm$0.4\% \\
\hline

\end{tabular}

}
\end{center}

\end{table*}

\begin{table*}
\caption{Mean absolute error of NNRank, SVM, GP-MAP, and GP-EP on the eight datasets. 
SVM denotes the support vector machine method (Shashua \& Levin, 2003; Chu \& Ghahramani, 2005a). 
GP-MAP and GP-EP are two Gaussian process methods using Laplace approximation and expectation propagation 
respectively (Chu \& Ghahramani, 2005a).
The results are the average error over
20 trials along with the standard deviation.
We use boldface to denote the best results.}
\label{absolute-error}

\begin{center}
{\footnotesize

\begin{tabular}{|c|c|c|c|c|}
\hline
Data & NNRank & SVM & GP-MAP & GP-EP \\ \hline
Triazines & 0.730$\pm$0.07  & 0.698$\pm$0.03 & \textbf{0.687$\pm$0.02} & 0.688$\pm$0.03\\
Pyrimidines &0.450$\pm$0.10  & 0.450$\pm$0.11 & 0.427$\pm$0.09 & \textbf{0.392$\pm$0.07}\\
Diabetes & \textbf{0.546$\pm$0.15}  & 0.746$\pm$0.14 & 0.662$\pm$0.14 & 0.665$\pm$0.14 \\
%Wisconsin &1.08$\pm$0.094  & \textbf{1.0031+0.0727} & 1.0102$\pm$0.0937 & 1.0141$\pm$0.0932\\
Machine &0.186$\pm$0.04 & 0.192$\pm$0.04 & \textbf{0.185$\pm$0.04} & 0.186$\pm$0.04\\
Auto MPG &0.281$\pm$0.02  & 0.260$\pm$0.02 & 0.241$\pm$0.02 & \textbf{0.241$\pm$0.02}\\
Boston & 0.295$\pm$0.04  & 0.267$\pm$0.02 & 0.260$\pm$0.02 & \textbf{0.259$\pm$0.02} \\
Stocks & 0.127$\pm$0.02   & \textbf{0.108$\pm$0.02} & 0.120$\pm$0.02 & 0.120$\pm$0.02 \\
Abalone &\textbf{0.226$\pm$0.01}  & 0.229$\pm$0.01 & 0.232$\pm$0.01 & 0.234$\pm$0.01 \\
\hline
\end{tabular}

}
\end{center}
\end{table*}

\section{Discussion and Future Work}
We have described a simple yet novel approach to adapt traditional neural networks 
for ordinal regression. 
Our neural network approach can be considered a generalization of one-layer perceptron
approach (Crammer \& Singer, 2002) into multi-layer. 
On the standard benchmark of ordinal regression, our method outperforms 
standard neural networks used for classification. Furthermore, on the same benchmark, our method achieves
the similar performance as the two state-of-the-art methods 
(support vector machines and Gaussian processes) for ordinal regression.

Compared with existing methods for ordinal regression, 
our method has several advantages of neural networks. 
First, like the perceptron approach (Crammer \& Singer, 2002), our method can learn in both batch and online mode. 
The online learning ability 
makes our method a good tool for adaptive learning in the real-time. The multi-layer structure of 
neural network and the non-linear transfer function 
give our method the stronger fitting ability than perceptron methods. 

Second, the neural network can be trained on very large datasets iteratively, while training
is more complex than support vector machines and Gaussian processes.
Since the training process of our method is the same as traditional neural networks,
average neural network users can 
use this method for their tasks. 

Third, neural network method can make rapid prediction once models are trained. 
The ability of learning on very large dataset and predicting in time 
makes our method a useful and competitive tool for ordinal regression tasks,
particularly for time-critical and large-scale ranking problems 
in information retrieval, web page ranking, collaborative filtering, and
 the emerging fields of Bioinformatics. We are currently applying the method to 
rank proteins according to their structural relevance with respect to a query protein (Cheng \& Baldi, 2006). 
To facilitate the application of this new approach,
we make both NNRank and NNClass to accept a general input format and freely available
at http://www.eecs.ucf.edu/$^\sim$jcheng/cheng\_software.html.

There are some directions to further improve the neural network (or multi-layer perceptron) approach for 
ordinal regression. One direction is to design a transfer function
to ensure the monotonic decrease of the outputs of the neural network; the 
other direction is to 
derive the general error bounds of the method under the binary classification
framework (Li \& Lin, 2006). Furthermore, the other flavors of implementations of the multi-threshold multi-layer 
perceptron approach for ordinal regression are possible. 
Since machine learning ranking is a fundamental problem that has wide applications in many diverse domains such 
as web page ranking, information retrieval, image retrieval, collaborative filtering, bioinformatics and so on, we believe
the further exploration of the neural network (or multi-layer perceptron) approach 
for ranking and ordinal regression is worthwhile.
%directly
%generalizing the perceptron approach \cite{crammer-singer-02}
%can be applied to ordinal regression.   

\bibliographystyle{mlapa}

\bibliography{rank}

\end{document}